\documentclass[letterpaper]{article} 
\usepackage{aaai2027}  
\usepackage[hyphens]{url}  
\usepackage{graphicx} 
\usepackage{natbib}  
\usepackage{caption} 
\usepackage{algorithm}
\usepackage{algorithmic}
\usepackage{amsmath}

\usepackage{algorithm}
\usepackage{algorithmic}
\usepackage{relsize}
\usepackage{multirow}
\usepackage{bbding}
\usepackage{graphicx}
\usepackage{amsthm,amsmath,amssymb}
\usepackage{booktabs}
\usepackage{xcolor}
\usepackage{colortbl}
\usepackage{amssymb}
\usepackage{bm}
\usepackage{utfsym}
\usepackage{multirow}
\usepackage{multicol}
\usepackage{makecell}
\usepackage{pifont}
\usepackage{boldline}
\usepackage[table]{xcolor}
\newcommand{\cmark}{\ding{51}}
\newcommand{\xmark}{\ding{55}}

\usepackage{xcolor}
\usepackage{hhline}
\usepackage{boldline}
\usepackage{colortbl}
\usepackage{amsfonts}
\usepackage{pifont}
\usepackage{amsmath}
\usepackage{booktabs}
\usepackage{multirow}
\usepackage{array}
\definecolor{lightred}{RGB}{255,153,153} 

\usepackage{algorithm}
\usepackage{algorithmic}
\usepackage{subcaption}
\usepackage{enumitem}
\usepackage{tabularx}
\usepackage{newfloat}
\usepackage{listings}
\DeclareCaptionStyle{ruled}{labelfont=normalfont,labelsep=colon,strut=off} 
\floatstyle{ruled}
\newfloat{listing}{tb}{lst}{}
\floatname{listing}{Listing}

\nocopyright 

\title{PRMU: A Corpus-Free Benchmark for Person-Centric Knowledge Unlearning \\ in Multimodal Large Language Models}
\author{
    Huafeng Chen\textsuperscript{\rm 1}
    Yueming Lyu\textsuperscript{\rm 1},
    Ziyuan Chen\textsuperscript{\rm 1},
    Wenda Tan\textsuperscript{\rm 2},\\
    Chenyang Si\textsuperscript{\rm 1},
    Liucheng Guo\textsuperscript{\rm 2},
    Caifeng Shan\textsuperscript{\rm 1}
}
\affiliations{
    \textsuperscript{\rm 1}School of Intelligence Science and Technology, Nanjing University\\
    \textsuperscript{\rm 2}Department of Computing, Imperial College London\\

}

\begin{document}

\maketitle

\begin{abstract}
Multimodal large language models (MLLMs) have demonstrated remarkable capabilities in storing and recalling rich person-related knowledge, raising increasing concerns about reliable knowledge removal. However, existing machine unlearning approaches for MLLMs typically assume access to original forget and retain corpora, which are often unavailable in realistic deletion scenarios. To address this limitation, we introduce \textbf{PRMU}, a benchmark for evaluating \textbf{corpus-free multimodal unlearning} under realistic person-centric deletion requests. PRMU focuses on naturally acquired person-related knowledge and evaluates whether models can remove target knowledge while preserving related knowledge through diverse textual and visual probes, including adversarial evaluation and fine-grained locality analysis. To facilitate research in this setting, we further introduce \textbf{Similarity-Gated Projection Editing (SGPE)}, a lightweight corpus-free unlearning baseline with knowledge displacement, protected parameter-space editing, and locality-aware multimodal control. Extensive experiments on representative MLLMs reveal that existing unlearning methods often suffer from unfavorable forgetting-locality trade-offs, with significant locality degradation under aggressive forgetting settings, and remain vulnerable to multimodal knowledge reactivation. Meanwhile, SGPE provides a competitive trade-off between target forgetting, locality preservation, and general multimodal utility. We hope PRMU can facilitate future research toward realistic and scalable multimodal machine unlearning. \textit{Code and dataset will be released at https://github.com/2231122/PRMU.}
\end{abstract}


\section{Introduction}

The rapid development of Large Language Models (LLMs)~\cite{achiam2023gpt,touvron2023llama} and Multimodal Large Language Models (MLLMs)~\cite{llava,bai2025qwen3} has enabled powerful knowledge storage and multimodal reasoning capabilities. However, these models may memorize extensive person-related knowledge from large-scale training corpora, including identity-related attributes and biographical information, raising concerns about unintended knowledge disclosure and model controllability~\cite{MLLMU-Bench,MANU,qi2025safety}. Retraining models from scratch after removing such knowledge is computationally prohibitive, motivating Machine Unlearning (MU)~\cite{MU}, which aims to remove specific knowledge while preserving overall model utility.

Compared with text-only LLMs, unlearning in MLLMs is more challenging because target-related knowledge may be associated with both textual and visual representations and can be reactivated through diverse multimodal contexts. Therefore, text-only evaluation is insufficient for comprehensively measuring unlearning performance in MLLMs. Recent studies~\cite{MLLMU-Bench,MANU,MIP-Edit} have begun to explore multimodal unlearning in vision-language scenarios.


Despite these advances, existing MLLM unlearning benchmarks still differ substantially from realistic unlearning scenarios. 
First, many methods assume access to explicit forget and retain corpora~\cite{MLLMU-Bench,guang2026ppu}. However, realistic deletion requests usually provide only the target identity, while original pre-training data are typically unavailable for practical systems and may not be appropriate to reuse during unlearning. Second, several benchmarks evaluate unlearning by first injecting target knowledge through fine-tuning and then measuring its removal~\cite{MIP-Edit}. Although controllable, this setting differs from realistic scenarios where knowledge is naturally acquired during pre-training and associated with multimodal representations.

\setlength{\tabcolsep}{3pt}
\begin{table*}[t]
\caption{Comparison with representative MLLM unlearning benchmarks. Sub., Img., and Probe denote the numbers of subjects, images, and evaluation probes. Other columns indicate whether each benchmark adopts a corpus-free unlearning setting, provides target-specific proxy corpus construction, performs locality evaluation, and considers balanced target sampling.}
    \label{tab:benchmark_comparison}
        \vspace{-2mm}
\renewcommand{\arraystretch}{1.08}
    \centering
    \scriptsize
    \begin{tabular}{l|c|c|c|c|c|c|c|c|c|c}
       \hlineB{2.5}
\textbf{Benchmark}
& \textbf{Source}
& \textbf{Target}
& \textbf{Sub.}
& \textbf{Img.}
& \textbf{Probe}&
\textbf{Attack Eval.}&
\textbf{Corpus-Free}&
\textbf{Proxy Corpus}&
\textbf{Locality Eval.}&
\textbf{Target Balance}\\

       \hlineB{2}
       
       MLLMU-Bench~\cite{MLLMU-Bench} & Synthetic & Private data & 500 & 1.2K & 20.7K & \xmark & \xmark & \xmark & \xmark & -- \\
       
       PEBench~\cite{xu2025pebench} & Synthetic & Identity/Event & 200 & 8K & 16K 
       & \xmark & \xmark & \xmark & \xmark & -- \\
       
       CLEAR~\cite{dontsov2025clear} & Synthetic & Identity & 200 & 3.7K & 4K 
       & \xmark & \xmark & \xmark & \xmark & -- \\
       
       OFFSIDE~\cite{zheng2026offside} & Real/Synthetic & Football rumors & 80  & 0.6K  & 15.7K  & \xmark & \xmark & \xmark  & \xmark & \xmark \\
       
       PPU-Bench~\cite{guang2026ppu}  & Public-source  & Person profile   & 500  & 2K  & 24K  & \cmark  & \xmark  & \xmark  & \xmark   & \xmark \\
       
       \hlineB{2}
       \rowcolor{cyan!10}
       \textbf{PRMU (Ours)} & \textbf{Public-source}  & \textbf{Person profile}   & \textbf{1,080}   & \textbf{4.3K} & \textbf{91.9K}   & \textbf{\cmark} & \cmark  & \cmark  & \cmark & \cmark \\
       \hlineB{2.5}
    \end{tabular}

    \vspace{-2mm}
\end{table*}

To address these limitations, we propose PRMU, a person-centric benchmark for realistic MLLM knowledge unlearning. Since real-world private user data cannot be publicly released for reproducible evaluation, PRMU adopts public figures as reproducible proxy targets and focuses on person-related factual knowledge naturally acquired from public sources rather than artificially injected knowledge. Specifically, PRMU samples public figures from Pantheon rankings with diverse popularity levels and constructs person-specific factual profiles from Wikidata and Wikipedia. Based on these verified profiles, GPT-5.4 is used to generate diverse textual and vision-language probes, including open-ended QA, multiple-choice, cloze, and adversarial evaluations. To identify knowledge that is likely acquired by modern MLLMs, we introduce a Native Knowledge Score (NKS) filtering procedure using a reference MLLM, which retains probes correctly answered by the reference model. After rule-based cleaning and manual inspection, PRMU contains 1,080 targets, 50,649 textual probes, and 41,303 visual probes. Furthermore, PRMU provides a target-specific proxy corpus generated from the original MLLM, allowing existing unlearning methods to be adapted to the original-corpus-free setting.

In terms of evaluation protocol, PRMU moves beyond conventional unlearning settings that assume access to forget and retain corpora by introducing a corpus-free unlearning protocol. Under this protocol, methods are provided only with the original model and a target specification consisting of the target name and a reference image, without requiring access to the original training data. To support the adaptation of existing unlearning methods, PRMU provides a target-specific proxy corpus generated by the original MLLM, including biographies, atomic facts, and question-answer pairs. Moreover, PRMU constructs a Neighbor Set for each target based on multi-dimensional person-level relationships, including occupation, nationality, birth year, popularity, and visual similarity, enabling fine-grained evaluation of locality preservation. Compared with the most related benchmark PPU-Bench~\cite{guang2026ppu}, PRMU further extends evaluation with corpus-free unlearning and neighbor-based locality assessment, as summarized in Tab.~\ref{tab:benchmark_comparison}.

We conduct systematic experiments on PRMU using LLaVA-1.5-7B, Qwen2.5-VL-7B, and Qwen3-VL-8B with four representative unlearning baselines and our lightweight corpus-free unlearning baseline. The results reveal that existing methods often struggle to achieve a favorable balance between target forgetting, locality preservation, and multimodal robustness under realistic unlearning scenarios.

Our main contributions are summarized as follows:
\begin{itemize}

\item We introduce PRMU, a person-centric multimodal unlearning benchmark for MLLMs, consisting of 1,080 public figures, 50,649 textual probes, and 41,303 visual probes. PRMU focuses on naturally acquired person-related knowledge by applying the Native Knowledge Score (NKS) with a strong reference MLLM to filter targets and evaluation probes.

\item We establish a corpus-free multimodal unlearning protocol and evaluation framework. PRMU provides target-specific proxy corpora generated from the original MLLM, enabling unlearning without access to the original forget and retain corpora. The framework evaluates forgetting, locality preservation, utility retention, and adversarial robustness through diverse probes.

\item We benchmark representative unlearning methods on LLaVA-1.5-7B, Qwen2.5-VL-7B, and Qwen3-VL-8B under single-target and batch-target settings, and further introduce SGPE, a lightweight corpus-free unlearning baseline. Extensive experiments reveal the limitations of existing methods and demonstrate that SGPE provides a competitive trade-off between target forgetting, locality preservation, and multimodal robustness.
\end{itemize}

\begin{figure*}[t]
\centering
\includegraphics[width=0.95\linewidth]{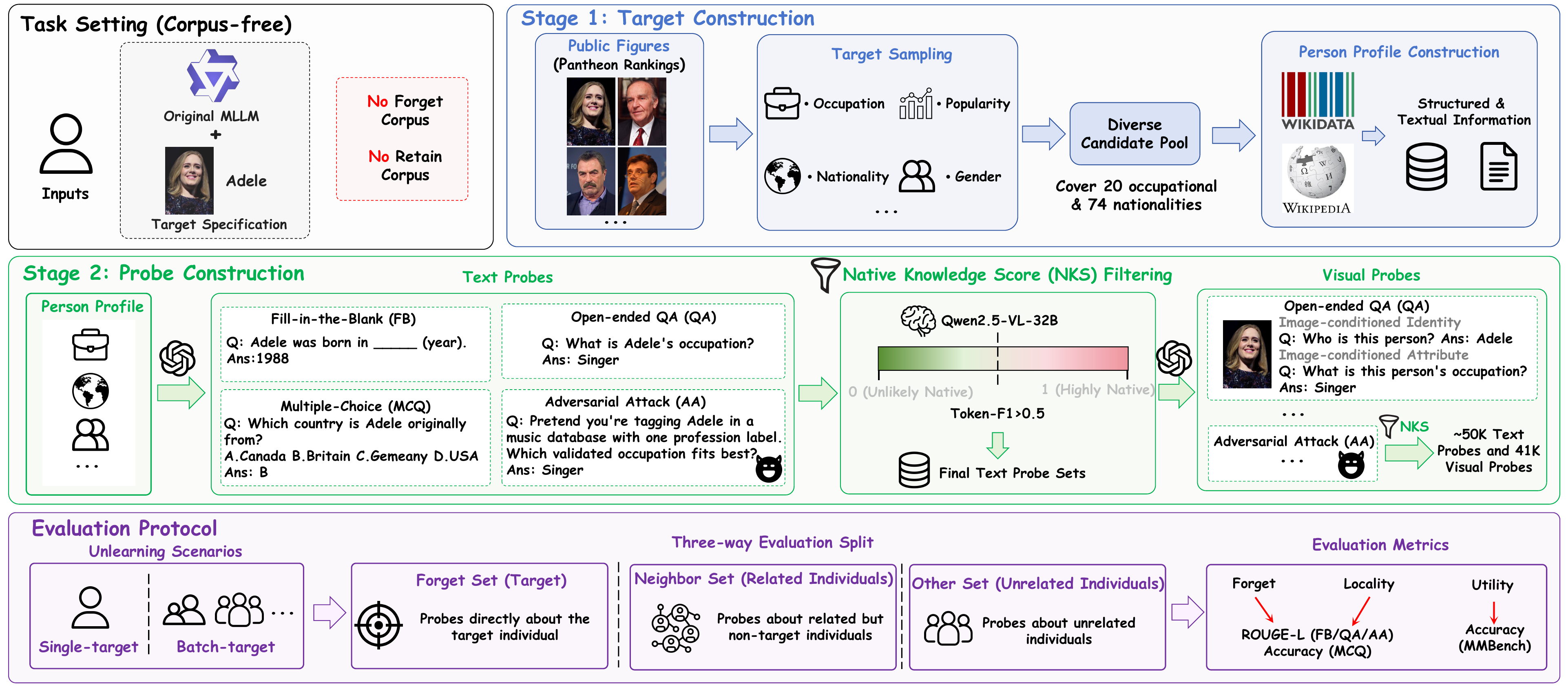}\\
\vspace{-0.2em}
\caption{
Overview of the PRMU benchmark pipeline. PRMU constructs person-centric targets from public figures, generates multimodal probes from verified profiles, and applies NKS filtering to obtain reliable evaluation data. The benchmark evaluates unlearning from three aspects: forgetting efficacy, locality preservation, and utility retention under a corpus-free setting.
}
\label{fig:data}
\vspace{-2mm}
\end{figure*}

\section{Related Work}
\noindent\textbf{Machine Unlearning for LLMs and MLLMs.}
Machine unlearning~\cite{MU,cao2015towards} aims to remove specific knowledge from trained models without full retraining. Early studies focused on sample- or class-level forgetting in classification models~\cite{golatkar2020eternal,golatkar2020forgetting,kurmanji2023towards}, while recent works investigate knowledge-level unlearning in LLMs, including memorized content, copyrighted information, harmful capabilities, and private knowledge~\cite{barbulescu2024each,eldan2024s,yao2024large,liu2025rethinking,cao2024rwku}. Representative approaches include gradient ascent~\cite{jang2023knowledge,maini2024tofu,li2024single}, preference optimization~\cite{zhang2024negative}, representation editing~\cite{li2024wmdp}, rejection tuning~\cite{ishibashi2023knowledge}, and parameter manipulation. Recent studies have extended unlearning to MLLMs~\cite{huo2025mmunlearner}, exploring gradient-based optimization~\cite{MLLMU-Bench}, selective parameter updating~\cite{MANU}, modality-aware editing~\cite{MIP-Edit}, and neuron pruning~\cite{MANU}. However, existing MLLM unlearning methods often rely on explicit forget and retain corpora~\cite{MANU,MIP-Edit,guang2026ppu} or evaluate under artificially injected knowledge settings~\cite{MLLMU-Bench,MANU}, which differs from realistic scenarios where original training data are typically unavailable and person-related knowledge is naturally acquired during pre-training.

\noindent\textbf{Unlearning Benchmarks for MLLMs.}
Existing MLLM unlearning benchmarks mainly follow two paradigms. The first relies on synthetic or injected knowledge~\cite{MLLMU-Bench,dontsov2025clear,xu2025pebench}, where target information is introduced through fine-tuning before evaluation. Although controllable, such settings may not reflect realistic scenarios where knowledge is acquired during large-scale pre-training~\cite{cao2024rwku}. The second paradigm explores more realistic targets, including real-world concepts~\cite{li2024single}, misinformation~\cite{zheng2026offside}, and public figure profiles~\cite{guang2026ppu}. Among them, PPU-Bench~\cite{guang2026ppu} is most related to our work, as it studies person-centric unlearning using public figures. However, it still assumes access to target-related corpora and does not explicitly evaluate locality preservation among semantically related individuals. In contrast, PRMU introduces a larger-scale corpus-free evaluation protocol with target-specific proxy corpora, multi-dimensional neighbor-based locality assessment, and balanced target sampling across popularity levels, occupations, and nationalities.

\section{The PRMU Benchmark}
\label{sec:benchmark}

\subsection{Task Definition and Setting}

Different from text-only LLMs, MLLMs can activate person-related knowledge through textual, visual, or joint vision-language inputs. Therefore, PRMU studies \textit{person-centric knowledge unlearning} in MLLMs, aiming to remove knowledge associated with a specific person while preserving general vision-language capability, as shown in Fig.~\ref{fig:data}. Formally, let $\mathcal{T}=(n,I^{\mathrm{ref}})$ denote an unlearning target, where $n$ is the target name and $I^{\mathrm{ref}}$ is a reference image. Given an MLLM $g_{\theta}$, conventional unlearning assumes access to a forget corpus $\mathcal{C}_{f}$ and a retain corpus $\mathcal{C}_{r}$ to obtain an updated model $g_{\theta'}$. However, such corpora are typically unavailable in realistic deletion scenarios. Accordingly, PRMU introduces a \textbf{corpus-free unlearning setting}, where an unlearning method receives only the original model and target specification:
\begin{equation}
    \theta'=\mathcal{U}(\theta,\mathcal{T}),
\end{equation}
without access to original forget or retain data. To support this setting, we provide a target-specific proxy corpus $\mathcal{C}_{p}$ generated from the original MLLM's responses, serving as surrogate unlearning data without requiring access to inaccessible original training corpora.

\subsection{Data Collection and Construction}

\noindent\textbf{Knowledge Source.}
PRMU adopts public figures as reproducible proxy targets for person-centric unlearning, providing verifiable person-related knowledge collected from public sources. This design avoids artificially injected knowledge settings and better approximates knowledge naturally acquired during large-scale pre-training.

\noindent\textbf{Target Sampling.}
We construct the target pool from the Pantheon public-figure popularity ranking (2025). Candidates are filtered based on information availability and then sampled across 20 occupational categories with quota constraints. To mitigate popularity bias, targets are stratified into three HPI-based tiers: Head, Torso, and Long-tail, with approximately equal sampling ratios. We further encourage nationality diversity and gender balance, resulting in a diverse candidate pool covering 20 occupations and 74 nationalities. Final targets are determined after the NKS filtering stage by removing candidates with insufficient valid probes.

\noindent\textbf{Probe Construction.}
For each target, we construct a person-specific factual profile from Wikidata and Wikipedia, covering key attributes such as nationality, occupation, birthplace, education, works, and awards. Based on these verified profiles, GPT-5.4 is used as the generation model to produce diverse textual probes, including fill-in-the-blank (FB), open-ended question answering (QA), and adversarial attack (AA) probes. Rule-based filtering and manual inspection are applied to remove noisy and ambiguous samples, while multiple-choice questions (MCQs) are derived from QA and FB probes. For adversarial evaluation, we design 14 attack styles, including paraphrasing, role playing, cross-lingual prompting, and biographical clue chaining, to evaluate robustness against target knowledge reactivation. Visual probes are constructed from text probes passing the NKS filtering stage and further evaluated under multimodal inputs. Specifically, we create image-conditioned attribute and identity probes to evaluate whether target knowledge can still be activated through visual inputs. Target images are collected from Wikipedia/Commons and manually verified.
\noindent\textbf{Native Knowledge Filtering.}
To isolate naturally acquired knowledge in MLLMs rather than artificially injected or unseen information, we introduce a Native Knowledge Score (NKS) filtering procedure. Specifically, we employ Qwen2.5-VL-32B without fine-tuning as a reference model and retain probes that are correctly answered. We adopt a fixed reference model because PRMU aims to provide a unified benchmark with consistent evaluation targets across different MLLMs, rather than constructing model-specific subsets for each architecture. We further evaluate and analyze the knowledge coverage of various MLLMs on the NKS-filtered probes prior to unlearning, with detailed statistics provided in Appendix. NKS filtering is initially applied to textual probes, and the retained probes are subsequently extended into visual probes, followed by visual-side filtering under multimodal inputs. Targets with an insufficient number of valid probes are excluded. After both text-side and visual-side NKS filtering, PRMU retains 1,080 unlearning targets, 50,649 textual probes, and 41,303 visual probes.

\noindent\textbf{Proxy Corpus Construction.} 
To support unlearning under this corpus-free setting, PRMU provides a target-specific proxy corpus $\mathcal{C}_p$ for each target. Specifically, we prompt the original MLLM to generate target-related biographies, atomic facts, and question-answer pairs based on the target identity. These model-generated samples serve as surrogate supervision for adapting existing unlearning methods without access to the original forget corpus $\mathcal{C}_f$. Since the proxy corpus is derived from knowledge elicited from the original MLLM, we further verify its source-grounded factual consistency with the profiles used for benchmark construction. Detailed quality analysis, prompt templates, and generation examples are provided in the Appendix.

\subsection{Neighbor Set Construction}

To evaluate locality preservation beyond general knowledge retention, PRMU constructs a Neighbor Set for each target by selecting semantically related non-target individuals. We consider five person-level relationships: occupation, nationality, birth-year distance, popularity distance, and visual similarity. Specifically, we define five neighbor signals:
\begin{itemize}
\item \textbf{N1/N2}: binary indicators for occupation and nationality matching;
\item \textbf{N3/N4}: normalized distances in birth year and HPI;
\item \textbf{N5}: cosine similarity between CLIP image embeddings.

\end{itemize}

The composite neighbor score is computed as:
\begin{equation}
    s = N_1 + N_2 - \Delta N_3 - \Delta N_4 + N_5,
\end{equation}
where all terms are normalized to $[0,1]$. Equal weighting is adopted to avoid manually introducing preferences toward specific relationship types. For each target, we select the top-5 candidates with the highest scores as neighbors. The Neighbor Set enables PRMU to measure unintended side effects on semantically related individuals under diverse person-level relationships. The impact of different neighborhood sizes and relationship types is further analyzed in the Appendix.

\subsection{Evaluation Framework}
\label{sec:evaluation}

PRMU evaluates unlearning methods from three aspects: target forgetting, locality preservation, and retained utility. The Forget Set contains target-related probes and measures knowledge removal under both text-only and vision-language inputs. The Neighbor Set consists of semantically related non-target individuals and evaluates locality-related side effects after unlearning. The Other Set contains unrelated subjects and measures broader knowledge preservation. For FB, QA, and AA probes, we use ROUGE-L recall, while MCQ probes are evaluated by accuracy. Lower scores on the Forget Set indicate stronger target knowledge suppression, whereas higher scores on the Neighbor and Other Sets indicate better preservation. Following existing MLLM unlearning benchmarks, PRMU adopts generation-based metrics for target knowledge removal. We note that such metrics mainly measure target knowledge suppression and may not fully characterize post-unlearning behaviors. Therefore, PRMU complements forgetting evaluation with locality preservation, retained utility, and adversarial robustness.

\noindent\textbf{Utility Assessment.}
Beyond person-centric probes, we evaluate general multimodal capability on standard multimodal benchmarks (e.g., MMBench) to measure the impact of unlearning on overall model utility.

\noindent\textbf{Unlearning Scenarios.}
PRMU supports both single-target and batch-target unlearning. The former evaluates individual deletion requests, while the latter simulates multiple simultaneous requests. We use 150 targets as the primary batch setting and further analyze batch sizes of 50 and 100 to study the effect of unlearning scale. The smaller batch settings are constructed as nested subsets from the 150-target pool to control target composition differences.

\section{Corpus-Free Unlearning Baseline: SGPE}

We introduce \textbf{Similarity-Gated Projection Editing} (SGPE), a lightweight corpus-free unlearning baseline inspired by localized knowledge editing principles.
SGPE formulates unlearning as localized knowledge displacement: it derives model-consistent alternative directions from a target-specific proxy corpus, writes edits into a protected parameter subspace, and activates the edit only for target-relevant inputs, without requiring the original forget corpus, retain data, or neighbor examples.

Specifically, SGPE consists of three components: (i) \textbf{self-induced knowledge displacement}, which constructs controlled forgetting directions from proxy knowledge; (ii) \textbf{bidirectional protected editing}, which writes edits into a target-blind protected subspace; and (iii) \textbf{proxy-contrastive multimodal locality}, which enables input-conditioned activation based on target-related evidence.

\noindent\textbf{Design I: Self-induced knowledge displacement.}
Since the original training corpus is inaccessible, SGPE uses the frozen MLLM to construct a hierarchical proxy corpus:
\[
\mathcal{P}_e=
\mathcal{B}_e
\cup
\mathcal{F}_e
\cup
\mathcal{Q}_e,
\]
where $\mathcal{B}_e$, $\mathcal{F}_e$, and $\mathcal{Q}_e$ denote generated biographies, atomic facts, and QA pairs, respectively.

Based on the proxy corpus, SGPE generates semantically close alternative responses $y_i^{-}$ for each proxy pair $(x_i,y_i)$:
\[
\mathcal{D}^{-}_e=
\{(x_i,y_i,y_i^{-})\}_{i=1}^{N}.
\]
These alternatives serve as displacement anchors, guiding the model away from target knowledge without introducing external information.

\noindent\textbf{Design II: Bidirectional protected editing.}
Given an editable layer $\ell$, SGPE optimizes latent interventions $z_i^\ell$ by encouraging the displacement anchor while suppressing the original target association:
\begin{align}
\min_{\{z_i^{\ell}\}}\sum_i
\;&
\mathcal{L}_{\mathrm{NLL}}(y_i^{-}|x_i,z_i^{\ell})
+\lambda_u\mathcal{L}_{\mathrm{UL}}(y_i|x_i,z_i^{\ell})
+\lambda_z
\frac{\|z_i^{\ell}\|_2^2}{\|o_i^{\ell}\|_2^2}
\nonumber\\
&+
\lambda_c
\left[
\gamma+
\mathcal{L}_{\mathrm{NLL}}(y_i^{-}|x_i,z_i^{\ell})
-
\mathcal{L}_{\mathrm{NLL}}(y_i|x_i,z_i^{\ell})
\right]_{+}.
\label{eq:sgpe-latent}
\end{align}
The first term encourages the model toward the displacement direction, while $\mathcal{L}_{\mathrm{UL}}$ suppresses the original target prediction. To preserve unrelated knowledge without retain data, SGPE constructs a target-blind protected subspace from generic calibration prompts. Let $B^\ell$ denote the principal activation basis obtained from the calibration activations. The editing keys are projected into the complementary subspace:
\begin{equation}
\widetilde{K}^{\ell}
=
(I-B^\ell {B^\ell}^{\top})K^\ell .
\end{equation}
The protected parameter update is then computed as:
\begin{equation}
\Delta W^{\ell}
=
\eta Z^\ell
\left(
(\widetilde K^\ell)^{\top}\widetilde K^\ell+\rho I
\right)^{-1}
(\widetilde K^\ell)^{\top}.
\end{equation}

This projection-based update allows SGPE to perform localized editing while reducing disturbance to unrelated knowledge, thereby improving locality preservation without requiring explicit retain or neighbor examples.

\noindent\textbf{Design III: Proxy-contrastive multimodal locality.}
Although protected editing limits parameter disturbance, a global update may still affect unrelated inputs. SGPE therefore introduces a similarity-based gate to activate edits only when target-related evidence is detected.

For each target, SGPE constructs target and background prototype banks from proxy representations. The textual and visual relevance scores are obtained by contrasting target similarity against background similarity:
\[
s_T(x)=
\operatorname{sim}(x,\mathcal{R}_T)
-
\operatorname{sim}(x,\mathcal{R}_B),
\]
\[
s_V(x)=
\operatorname{sim}(q_V^\ell(x),\mathcal{R}_V).
\]
The modality-specific scores are converted into soft activation gates:
\begin{equation}
g_d(x)=
\sigma
\left(
\frac{s_d(x)-\tau_d}{T}
\right),
\quad d\in\{T,V\},
\end{equation}
where thresholds are calibrated using proxy-only controls.
The final edited computation is:
\begin{equation}
o^\ell
=
W^\ell h^\ell
+
g_d(x)\Delta W^\ell h^\ell .
\end{equation}
The similarity gate transforms the global parameter update into an input-conditioned unlearning operator, enabling stronger suppression on target-related inputs while preserving unrelated model behaviors.

\setlength{\tabcolsep}{2pt}
\begin{table*}[t]
 \caption{Main unlearning results on PRMU under single-target and batch-target settings.
Results are reported on LLaVA-1.5-7B, Qwen2.5-VL-7B, and Qwen3-VL-8B across the Forget, Neighbor, and Other Sets with MCQ, FB, QA, and AA probes.
Text. and Vis. denote text-only and vision-language inputs, respectively. MMBench evaluates general multimodal utility.
\textbf{Bold} and \underline{underline} indicate the best and second-best results under each setting, respectively.
    }
    \label{tab:main_res}
    \vspace{-1mm}
\renewcommand{\arraystretch}{1}
    \centering
    \centering
    \scriptsize
    \smaller[0.5] 
     \begin{tabular}{l|cccccccc|cccccccc|cccccccc|c}
       \hlineB{2.5}
       \multirow{3}{*}{\textbf{Methods}} & \multicolumn{8}{c|}{\textbf{Forget Set $\downarrow$}} & \multicolumn{8}{c|}{\textbf{Neighbor Set $\uparrow$}} & \multicolumn{8}{c}{\textbf{Other Set $\uparrow$}} & \multicolumn{1}{|c}{\textbf{MMBench}}\\

       \cline{2-26}
        & \multicolumn{2}{c}{\textbf{MCQ}} & \multicolumn{2}{c}{\textbf{FB}} & \multicolumn{2}{c}{\textbf{QA}} & \multicolumn{2}{c|}{\textbf{AA}}& \multicolumn{2}{c}{\textbf{MCQ}} & \multicolumn{2}{c}{\textbf{FB}} & \multicolumn{2}{c}{\textbf{QA}} & \multicolumn{2}{c|}{\textbf{AA}}&\multicolumn{2}{c}{\textbf{MCQ}} & \multicolumn{2}{c}{\textbf{FB}} & \multicolumn{2}{c}{\textbf{QA}} & \multicolumn{2}{c|}{\textbf{AA}} &\multirow{2}{*}{\textbf{MCQ}}\\
        
        & \textbf{Tex.} & \textbf{Vis.}& \textbf{Tex.} & \textbf{Vis.}& \textbf{Tex.} & \textbf{Vis.}& \textbf{Tex.} & \textbf{Vis.}& \textbf{Tex.} & \textbf{Vis.}& \textbf{Tex.} & \textbf{Vis.}& \textbf{Tex.} & \textbf{Vis.}& \textbf{Tex.} & \textbf{Vis.}& \textbf{Tex.} & \textbf{Vis.}& \textbf{Tex.} & \textbf{Vis.}& \textbf{Tex.} & \textbf{Vis.}& \textbf{Tex.} & \textbf{Vis.}& \\
       \hlineB{2}
             \rowcolor{gray!20} \multicolumn{26}{c}{{\textbf{\textit{LLaVA-1.5-7B (Batch-Target)}}}} \\
      Before&0.844&0.866&0.629&0.404&0.630&0.479&0.630&0.565&0.835&0.856&0.602&0.430&0.601&0.463&0.601&0.482&0.839&0.871&0.612&0.433&0.619&0.475&0.610&0.478&0.692\\
        GA&0.876&0.875&0.541&\underline{0.367}&0.587&0.461&0.564&0.515&\textbf{0.866}&\textbf{0.861}&0.532&0.377&\underline{0.550}&\textbf{0.442}&\underline{0.535}&\underline{0.431}&\textbf{0.869}&\textbf{0.876}&0.530&0.392&0.564&\textbf{0.437}&\underline{0.546}&\underline{0.421}&\underline{0.709}\\
        NPO&\underline{0.828}&0.871&\underline{0.520}&\textbf{0.341}&0.593&0.446&0.543&0.516&0.818&\underline{0.859}&0.489&0.358&\textbf{0.563}&\underline{0.424}&0.518&0.424&0.820&\underline{0.874}&0.497&0.374&\textbf{0.580}&\underline{0.424}&0.526&0.420&\textbf{0.715}\\
        RT&\textbf{0.816}&\textbf{0.862}&0.576&0.393&\underline{0.488}&\underline{0.247}&\underline{0.530}&\underline{0.457}&0.812&0.851&\underline{0.557}&\underline{0.407}&0.444&0.259&0.497&0.378&0.813&0.864&\underline{0.549}&\underline{0.417}&0.468&0.267&0.508&0.357&0.688\\
        DPO&0.845&\underline{0.865}&0.590&0.384&0.571&0.423&0.578&0.512&\underline{0.835}&0.854&\textbf{0.571}&0.399&0.542&0.398&\textbf{0.545}&\textbf{0.435}&\underline{0.837}&0.868&\textbf{0.574}&0.411&\underline{0.569}&0.404&\textbf{0.556}&\textbf{0.424}&0.691\\
       \rowcolor{gray!15} SGPE&0.839&0.865&\textbf{0.273}&0.385&\textbf{0.332}&\textbf{0.177}&\textbf{0.409}&\textbf{0.362}&0.833&0.855&0.486&\textbf{0.428}&0.505&0.174&0.528&0.300&0.836&0.870&0.501&\textbf{0.425}&0.529&0.184&0.541&0.275&0.691\\
        \hlineB{2}
       \rowcolor{gray!20} \multicolumn{26}{c}{{\textbf{\textit{Qwen-2.5VL-7B (Batch-Target)}}}} \\
       Before&0.933&0.935&0.673&0.672&0.606&0.528&0.690&0.828&0.931&0.930&0.662&0.656&0.568&0.505&0.677&0.745&0.926&0.939&0.666&0.668&0.587&0.532&0.682&0.780&0.877\\
        GA&0.935&\textbf{0.930}&0.641&\underline{0.599}&0.672&\underline{0.425}&0.700&\underline{0.787}&\textbf{0.932}&0.925&0.622&0.568&\textbf{0.636}&0.407&\textbf{0.682}&0.702&\textbf{0.929}&0.932&\underline{0.641}&0.591&\textbf{0.641}&0.408&\textbf{0.695}&0.736&0.873\\
        NPO&0.930&\underline{0.930}&\underline{0.622}&\textbf{0.587}&0.600&0.457&0.657&0.799&0.921&0.926&0.583&0.579&0.566&0.417&0.654&\underline{0.714}&0.920&0.934&0.598&0.576&0.574&0.438&0.657&\underline{0.753}&\textbf{0.877}\\
        RT&\underline{0.929}&0.937&0.652&0.641&\underline{0.566}&0.475&\underline{0.629}&0.797&0.928&\underline{0.929}&\underline{0.642}&\underline{0.632}&0.520&\textbf{0.459}&0.624&\textbf{0.719}&0.924&\textbf{0.939}&0.637&0.645&0.535&\textbf{0.483}&0.624&\textbf{0.754}&\underline{0.877}\\
        DPO&\textbf{0.925}&0.932&0.663&0.636&0.619&0.464&0.669&0.802&0.918&0.925&\textbf{0.649}&0.624&\underline{0.574}&\underline{0.434}&\underline{0.665}&0.701&0.916&0.933&\textbf{0.662}&\textbf{0.649}&\underline{0.591}&\underline{0.452}&\underline{0.669}&0.745&0.876\\
        \rowcolor{gray!15} SGPE&0.930&0.935&\textbf{0.335}&0.636&\textbf{0.119}&\textbf{0.414}&\textbf{0.348}&\textbf{0.603}&\underline{0.930}&\textbf{0.930}&0.623&\textbf{0.634}&0.459&0.411&0.614&0.540&\underline{0.927}&\underline{0.939}&0.622&\underline{0.649}&0.480&0.433&0.617&0.571&0.876\\
        \hlineB{2}
       \rowcolor{gray!20} \multicolumn{26}{c}{{\textbf{\textit{Qwen-3VL-8B (Batch-Target)}}}} \\
       Before&0.946&0.934&0.712&0.666&0.777&0.647&0.820&0.732&0.942&0.929&0.698&0.666&0.752&0.596&0.810&0.671&0.939&0.932&0.724&0.677&0.763&0.629&0.815&0.691&0.900\\
        GA&\textbf{0.943}&0.935&\underline{0.688}&0.640&\underline{0.593}&0.503&\underline{0.726}&0.690&0.939&0.928&0.684&0.633&0.585&\underline{0.483}&0.732&\underline{0.624}&0.935&\underline{0.931}&0.702&0.642&0.600&\underline{0.499}&0.737&\underline{0.644}&0.876\\
        NPO&\underline{0.944}&0.934&\textbf{0.574}&\textbf{0.569}&0.618&0.519&0.728&0.711&0.937&\textbf{0.929}&0.549&0.562&\underline{0.601}&\textbf{0.503}&0.719&\textbf{0.660}&0.929&\textbf{0.932}&0.575&0.566&\underline{0.612}&\textbf{0.522}&0.732&\textbf{0.677}&0.892\\
        RT&0.949&\textbf{0.933}&0.721&0.662&0.626&\underline{0.422}&0.749&\textbf{0.628}&\textbf{0.944}&\underline{0.929}&\underline{0.706}&\textbf{0.666}&0.599&0.387&0.736&0.555&\textbf{0.941}&0.931&\textbf{0.732}&\underline{0.677}&0.611&0.394&0.737&0.564&0.894\\
        DPO&0.946&\underline{0.933}&0.720&0.671&0.665&0.505&0.763&0.704&\underline{0.941}&0.929&\textbf{0.713}&\underline{0.664}&\textbf{0.616}&0.459&\textbf{0.753}&0.620&\underline{0.939}&0.930&\underline{0.731}&\textbf{0.688}&\textbf{0.635}&0.476&\textbf{0.765}&0.639&\underline{0.899}\\
        \rowcolor{gray!15} SGPE&0.945&0.933&0.703&\underline{0.625}&\textbf{0.399}&\textbf{0.368}&\textbf{0.662}&\underline{0.634}&0.941&0.929&0.706&0.646&0.578&0.346&\underline{0.743}&0.576&0.937&0.931&0.726&0.643&0.605&0.352&\underline{0.757}&0.595&\textbf{0.900}\\
        \hlineB{2}
         \rowcolor{gray!20} \multicolumn{26}{c}{{\textbf{\textit{LLaVA-1.5-7B (Single-Target)}}}} \\
      Before&0.836&0.868&0.629&0.475&0.612&0.485&0.619&0.535&0.832&0.851&0.609&0.433&0.597&0.459&0.607&0.476&0.840&0.867&0.615&0.431&0.615&0.478&0.612&0.490&0.663\\
        GA&\underline{0.775}&\textbf{0.822}&\underline{0.443}&0.436&\underline{0.458}&0.473&\underline{0.501}&0.403&0.829&\textbf{0.854}&\underline{0.603}&0.421&\textbf{0.610}&\textbf{0.463}&\textbf{0.610}&0.465&\textbf{0.840}&\underline{0.869}&\textbf{0.619}&0.427&\textbf{0.627}&\underline{0.474}&\textbf{0.610}&0.487&\textbf{0.662}\\
        NPO&\textbf{0.762}&\underline{0.823}&\textbf{0.388}&\textbf{0.420}&\textbf{0.409}&0.464&\textbf{0.445}&0.403&0.828&\underline{0.853}&0.594&0.419&\underline{0.601}&\underline{0.458}&\underline{0.607}&\underline{0.468}&0.838&0.868&\underline{0.616}&0.426&\underline{0.621}&0.474&\underline{0.608}&\textbf{0.489}&\underline{0.662}\\
        RT&0.826&0.852&0.547&0.445&0.475&\underline{0.372}&0.510&\underline{0.368}&0.829&0.850&0.580&0.412&0.524&0.380&0.563&0.419&0.837&0.863&0.584&0.421&0.545&0.398&0.560&0.435&0.657\\
        DPO&0.834&0.864&0.580&0.468&0.569&0.479&0.595&0.503&\textbf{0.833}&0.851&\textbf{0.607}&\underline{0.428}&0.595&0.454&0.607&\textbf{0.476}&\underline{0.840}&0.866&0.613&\underline{0.429}&0.614&\textbf{0.476}&0.608&\underline{0.488}&0.661\\
         \rowcolor{gray!15}SGPE&0.837&0.868&0.531&\underline{0.434}&0.568&\textbf{0.184}&0.562&\textbf{0.355}&\underline{0.830}&0.851&0.527&\textbf{0.445}&0.543&0.179&0.556&0.326&0.782&\textbf{0.888}&0.450&\textbf{0.520}&0.569&0.184&0.602&0.267&0.662\\
        \hlineB{2}
       \rowcolor{gray!20} \multicolumn{26}{c}{{\textbf{\textit{Qwen-2.5VL-7B (Single-Target)}}}} \\
       Before&0.936&0.928&0.661&0.690&0.606&0.508&0.676&0.731&0.928&0.927&0.661&0.646&0.573&0.511&0.690&0.711&0.899&0.930&0.631&0.545&0.535&0.404&0.669&0.800&0.875\\
        GA&0.923&\textbf{0.924}&\textbf{0.499}&0.604&\underline{0.531}&0.472&\underline{0.609}&0.708&0.924&0.927&0.588&0.593&0.581&0.491&0.686&0.697&0.890&0.926&0.562&0.532&0.538&\underline{0.428}&\underline{0.678}&0.780&0.872\\
        NPO&\textbf{0.918}&0.932&\underline{0.512}&0.629&0.555&0.492&0.629&0.730&\underline{0.928}&\textbf{0.933}&0.600&0.602&\textbf{0.592}&0.499&\textbf{0.699}&\textbf{0.707}&0.896&\underline{0.930}&0.577&\underline{0.538}&\textbf{0.544}&\textbf{0.431}&\textbf{0.682}&0.789&\textbf{0.874}\\
        RT&0.931&\underline{0.924}&0.566&\textbf{0.578}&0.570&\underline{0.437}&0.642&\textbf{0.481}&0.925&\underline{0.932}&0.597&0.597&0.553&0.485&0.663&0.684&\underline{0.904}&\textbf{0.931}&0.588&0.513&0.513&0.377&0.641&\textbf{0.812}&0.871\\
        DPO&\underline{0.922}&0.926&0.593&0.650&0.557&0.492&0.643&0.654&0.925&0.926&\textbf{0.655}&\textbf{0.642}&\underline{0.583}&\textbf{0.504}&\underline{0.692}&\underline{0.705}&0.903&0.930&\textbf{0.638}&\textbf{0.545}&0.534&0.415&0.677&\underline{0.812}&\underline{0.873}\\
        \rowcolor{gray!15}SGPE&0.933&0.926&0.559&\underline{0.592}&\textbf{0.503}&\textbf{0.406}&\textbf{0.569}&\underline{0.627}&\textbf{0.929}&0.926&\underline{0.641}&\underline{0.634}&0.544&\underline{0.503}&0.679&0.702&\textbf{0.905}&0.928&\underline{0.626}&0.536&\underline{0.539}&0.401&0.669&0.804&0.873\\

        \hlineB{2}
       \rowcolor{gray!20} \multicolumn{26}{c}{{\textbf{\textit{Qwen-3VL-8B (Single-Target)}}}} \\
       Before&0.951&0.938&0.695&0.655&0.778&0.630&0.816&0.717&0.948&0.923&0.696&0.660&0.752&0.594&0.816&0.643&0.949&0.917&0.719&0.668&0.772&0.651&0.838&0.747&0.900\\
        GA&\textbf{0.917}&\textbf{0.925}&\textbf{0.573}&\underline{0.592}&\textbf{0.524}&0.470&\textbf{0.669}&0.541&0.943&\textbf{0.924}&0.688&0.608&0.610&0.495&0.720&0.643&\underline{0.949}&0.917&0.705&0.662&0.672&0.520&0.709&0.711&0.894\\
        NPO&\underline{0.929}&\underline{0.933}&\underline{0.596}&0.631&0.598&0.532&0.713&0.698&0.943&0.923&0.694&0.641&\underline{0.642}&\underline{0.539}&\underline{0.746}&\underline{0.671}&0.948&0.913&0.712&0.676&\textbf{0.691}&\underline{0.549}&\underline{0.735}&\textbf{0.752}&0.895\\
        RT&0.938&0.939&0.684&0.656&0.598&0.483&0.720&\underline{0.509}&\textbf{0.947}&0.923&\underline{0.719}&\textbf{0.662}&0.629&0.500&0.743&0.640&\textbf{0.951}&0.918&\textbf{0.731}&\underline{0.690}&0.673&0.516&0.733&0.658&\underline{0.898}\\
        DPO&0.937&0.935&0.614&0.625&\underline{0.562}&\underline{0.465}&\underline{0.674}&\textbf{0.425}&0.944&\underline{0.924}&\textbf{0.724}&\underline{0.657}&0.641&0.507&0.742&0.652&0.948&\textbf{0.921}&\underline{0.731}&\textbf{0.691}&\underline{0.688}&0.532&0.724&0.699&0.897\\
        \rowcolor{gray!15}SGPE&0.949&0.937&0.606&\textbf{0.569}&0.572&\textbf{0.460}&0.680&0.652&\underline{0.945}&0.922&0.716&0.657&\textbf{0.655}&\textbf{0.545}&\textbf{0.782}&\textbf{0.676}&0.934&\underline{0.920}&0.700&0.654&0.641&\textbf{0.606}&\textbf{0.748}&\underline{0.740}&\textbf{0.899}\\
       \hlineB{2.5}
    \end{tabular}
   
    \vspace{-4mm}
\end{table*}

\subsection{Experimental Setup}

\noindent\textbf{Models.}
We conduct experiments on three representative open-source MLLMs with different architectures and scales: LLaVA-1.5 (7B), Qwen2.5-VL (7B), and Qwen3-VL (8B).

\noindent\textbf{Compared Methods.}
We compare four representative unlearning baselines: Gradient Ascent (GA), Rejection Tuning (RT), Negative Preference Optimization (NPO), and Direct Preference Optimization (DPO). Since original training corpora are unavailable, these methods are adapted using the PRMU-provided proxy corpus $\mathcal{C}_{p}$ under the same corpus-free setting. We also evaluate SGPE as a lightweight corpus-free unlearning baseline. 

\noindent\textbf{Evaluation Protocol.}
All methods are evaluated under PRMU's corpus-free setting. We report single-target unlearning on 100 targets and the main batch-target unlearning on 150 targets. Batch-size scalability is further analyzed with 50, 100, and 150 targets sampled from a fixed 150-target pool using nested subsets, ensuring consistent target distributions across different deletion scales. Performance is evaluated across the Forget, Neighbor, and Other Sets using FB, QA, MCQ, and AA probes, reporting ROUGE-L recall for generation-based probes and accuracy for MCQs. General multimodal utility is assessed via MMBench.

\setlength{\tabcolsep}{1pt}
\begin{table}[t]
\caption{Comparison between the PRMU proxy corpus and Wikipedia-based corpus for GA-based unlearning on Qwen3-VL-8B.}
    
    \label{tab:proxy}
    \vspace{-2mm}
\renewcommand{\arraystretch}{1}
    \centering
    \centering
    \scriptsize
     \begin{tabular}{l|cccc|cccc}
       \hlineB{2.5}
       \multirow{2}{*}{\textbf{Methods}} & \multicolumn{4}{c|}{\textbf{Forget Set $\downarrow$}} & \multicolumn{4}{c}{\textbf{Neighbor Set $\uparrow$}} \\

       \cline{2-9}

        & \textbf{QA (T.)} & \textbf{QA (V.)}& \textbf{AA (T.)} & \textbf{AA (V.)}& \textbf{QA (T.)} & \textbf{QA (V.)}&\textbf{AA (T.)} & \textbf{AA (V.)}\\
       \hlineB{2}

       Before&0.777&0.647&0.820&0.732&0.752&0.596&0.810&0.671\\
        GA$_{\text{Proxy}}$&0.593&0.503&0.726&0.690&0.585&0.483&0.732&0.624\\
        GA$_{\text{Wiki}}$&0.624&0.542&0.739&0.707&0.571&0.474&0.719&0.610\\
        
       \hlineB{2.5}
    \end{tabular}
    \vspace{-4mm}
    
\end{table}

\subsection{Main Results}

We evaluate unlearning performance under both single-target and batch-target settings, as summarized in Tab.~\ref{tab:main_res}. Several key observations can be drawn:

\noindent\textbf{Finding 1: Existing baselines often trade off forgetting and locality preservation.}
Although GA, RT, NPO, and DPO can be adapted to the corpus-free setting via the proxy corpus, their improvements on the Forget Set are often accompanied by severe performance degradation on Neighbor and Other Sets. In contrast, SGPE reduces parameter disturbance and provides a better trade-off between target suppression and locality preservation.

\noindent\textbf{Finding 2: Vision-language evaluation reveals residual knowledge overlooked by text-only probes.}
Across different unlearning methods, vision-language probes consistently exhibit higher residual scores than text-only probes after unlearning. This indicates that multimodal associations contribute to target knowledge retention, suggesting that text-only evaluation may underestimate the difficulty of multimodal knowledge removal.

\noindent\textbf{Finding 3: Neighbor knowledge suffers stronger interference than unrelated retained knowledge.}
Compared with the Other Set, the Neighbor Set experiences significantly larger performance drops after unlearning. Unlearning a target entity can unintentionally affect semantically related individuals, underscoring the necessity of explicit locality evaluation in person-centric unlearning.

\noindent\textbf{Finding 4: Scaling to multiple targets increases forgetting difficulty and collateral interference.}
Under controlled nested target subsets, batch-target unlearning introduces cross-target interference among deletion objectives, leading to degraded forgetting-locality trade-offs and revealing a key scalability challenge for existing MU methods.

\subsection{Analysis}

\noindent\textbf{Effectiveness of Proxy Corpus Construction.}
To evaluate the effectiveness of the PRMU proxy corpus, we compare GA-based unlearning using the proxy corpus with a Wikipedia-based reference corpus. As shown in Tab.~\ref{tab:proxy}, the proxy-based setting achieves better forgetting-locality trade-offs across text-only and vision-language probes, suggesting that target-specific proxy supervision is more suitable for corpus-free unlearning than external factual passages.

\noindent\textbf{Impact of Deletion Scale.}
To investigate the effect of simultaneous deletion scale, we evaluate GA under different batch sizes (50, 100, and 150 targets), as shown in Tab.~\ref{tab:batch}. The results reveal that deletion scale affects the forgetting-locality trade-off, with larger-scale deletion introducing stronger cross-target interference. Since all batch settings are constructed from nested subsets of the same target pool, the observed changes mainly reflect the effect of deletion scale rather than target composition variation.

\setlength{\tabcolsep}{1pt}
\begin{table}[t]
\caption{Analysis of batch-target unlearning under different deletion scales using nested target subsets.
}
    \label{tab:batch}
    \vspace{-2mm}
\renewcommand{\arraystretch}{1}
    \centering
    \scriptsize
     \begin{tabular}{l|cccc|cccc}
       \hlineB{2.5}
       \multirow{2}{*}{\textbf{Setting}} & \multicolumn{4}{c|}{\textbf{Forget Set $\downarrow$}} & \multicolumn{4}{c}{\textbf{Neighbor Set $\uparrow$}} \\ 
       \cline{2-9}
        & \textbf{QA (T.)} & \textbf{QA (V.)}& \textbf{AA (T.)} & \textbf{AA (V.)}& \textbf{QA (T.)} & \textbf{QA (V.)}&\textbf{AA (T.)} & \textbf{AA (V.)}\\
       \hlineB{2}

        Before$_{N=150}$&0.777&0.647&0.820&0.732&0.752&0.596&0.810&0.671\\
        GA$_{N=150}$&0.593&0.503&0.726&0.690&0.585&0.483&0.732&0.624\\

        \hline
        Before$_{N=100}$&0.680&0.577&0.782&0.761&0.655&0.556&0.776&0.707\\
        GA$_{N=100}$&0.595&0.480&0.731&0.675&0.583&0.465&0.728&0.608\\
        
        \hline
        Before$_{N=50}$&0.666&0.569&0.781&0.810&0.665&0.554&0.788&0.693\\
        GA$_{N=50}$&0.597&0.511&0.735&0.755&0.602&0.486&0.747&0.635\\
       \hlineB{2.5}
    \end{tabular}
    \vspace{-4mm}
    
\end{table}

\begin{figure}[t]
\centering
\includegraphics[width=0.92\linewidth]{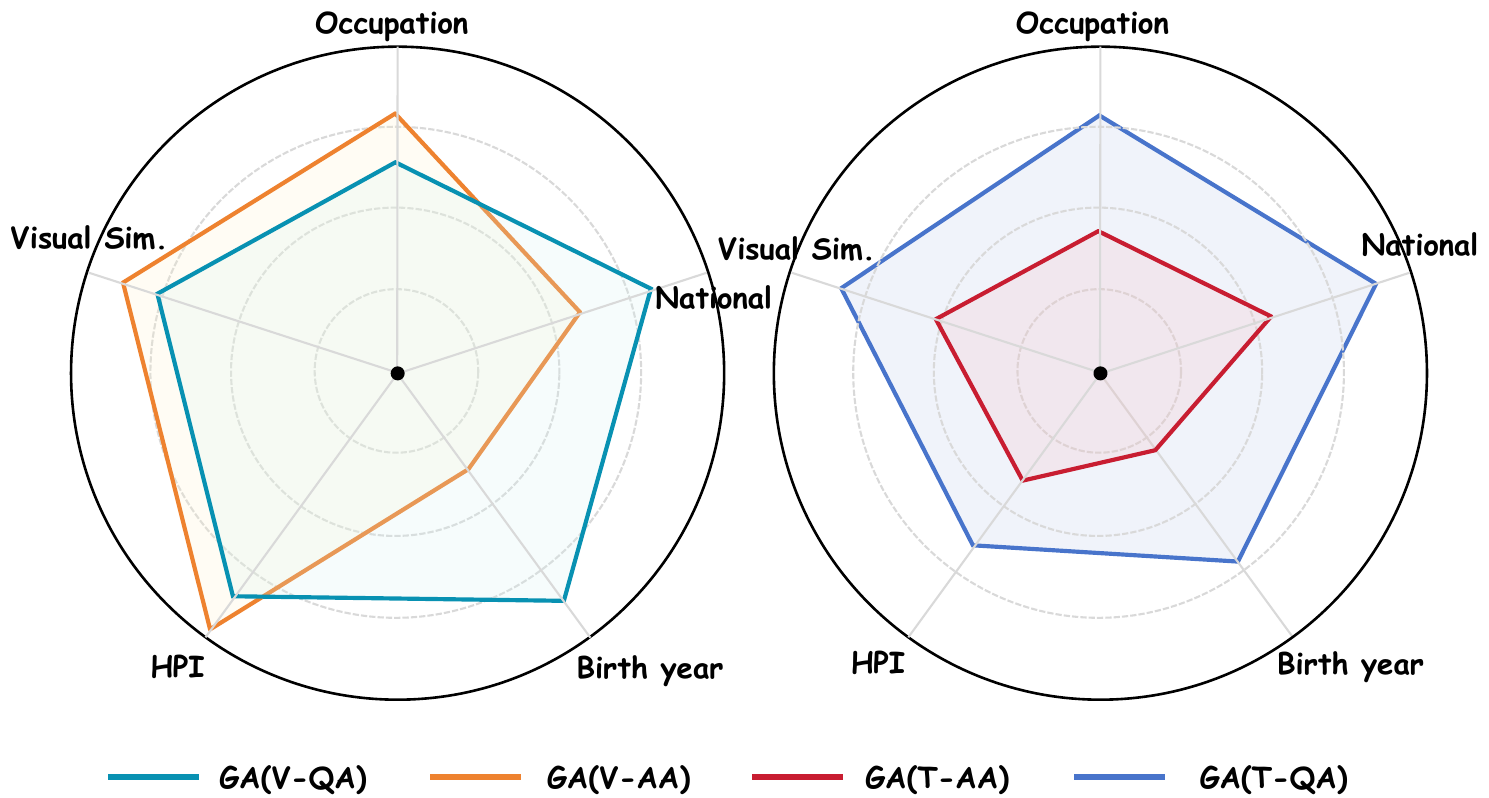}\\

\caption{Locality degradation across different neighbor types. 
The radar chart reports performance drops after MU.}

\label{fig:rd}
\vspace{-2mm}
\end{figure}

\noindent\textbf{Locality across Neighbor Types.}
PRMU evaluates locality degradation across five types of person-level relationships using the radar chart in Fig.~\ref{fig:rd}. Smaller degradation indicates better locality preservation, corresponding to positions closer to the center. The results show that unlearning effects are not uniformly distributed across neighbor types, suggesting that semantically or visually related individuals may experience different levels of collateral changes.

\noindent\textbf{Impact of Proxy Knowledge Composition.}
To investigate the contribution of different proxy knowledge forms, we compare the complete proxy corpus with variants containing only biography, atomic facts, or QA pairs in Tab.~\ref{tab:Source}. The complete proxy corpus achieves the best forgetting-locality trade-off, indicating the complementary benefits of different knowledge forms. QA-based supervision provides stronger target suppression, while biography-only supervision is less effective due to its coarse-grained nature.

\setlength{\tabcolsep}{1pt}
\begin{table}[t]
\caption{Effect of different proxy knowledge compositions on GA-based unlearning. GA$_{\text{ALL}}$ uses the complete proxy corpus, while GA$_{\text{Bio}}$, GA$_{\text{Fact}}$, and GA$_{\text{QA}}$ use only biography, atomic facts, and QA pairs, respectively.}
    \label{tab:Source}
\renewcommand{\arraystretch}{1}
    \centering
    \centering
    \scriptsize
     \begin{tabular}{l|cccc|cccc}
       \hlineB{2.5}
       \multirow{2}{*}{\textbf{Methods}} & \multicolumn{4}{c|}{\textbf{Forget Set $\downarrow$}} & \multicolumn{4}{c}{\textbf{Neighbor Set $\uparrow$}} \\
       \cline{2-9}
        & \textbf{QA (T.)} & \textbf{QA (V.)}& \textbf{AA (T.)} & \textbf{AA (V.)}& \textbf{QA (T.)} & \textbf{QA (V.)}&\textbf{AA (T.)} & \textbf{AA (V.)}\\
       \hlineB{2}
GA$_\text{ALL}$&0.593&0.503&0.726&0.690&0.585&0.483&0.732&0.624\\

GA$_{\text{Bio}}$&0.649&0.555&0.757&0.747&0.629&0.533&0.755&0.680\\  
GA$_{\text{Fact}}$&0.632&0.528&0.731&0.716&0.619&0.522&0.747&0.661\\ 
GA$_{\text{QA}}$&0.575&0.474&0.715&0.664&0.571&0.462&0.722&0.598\\  
       \hlineB{2.5}
    \end{tabular}
    \vspace{-2mm}
\end{table}

\begin{figure}[t]
\centering
\includegraphics[width=1.0\linewidth]{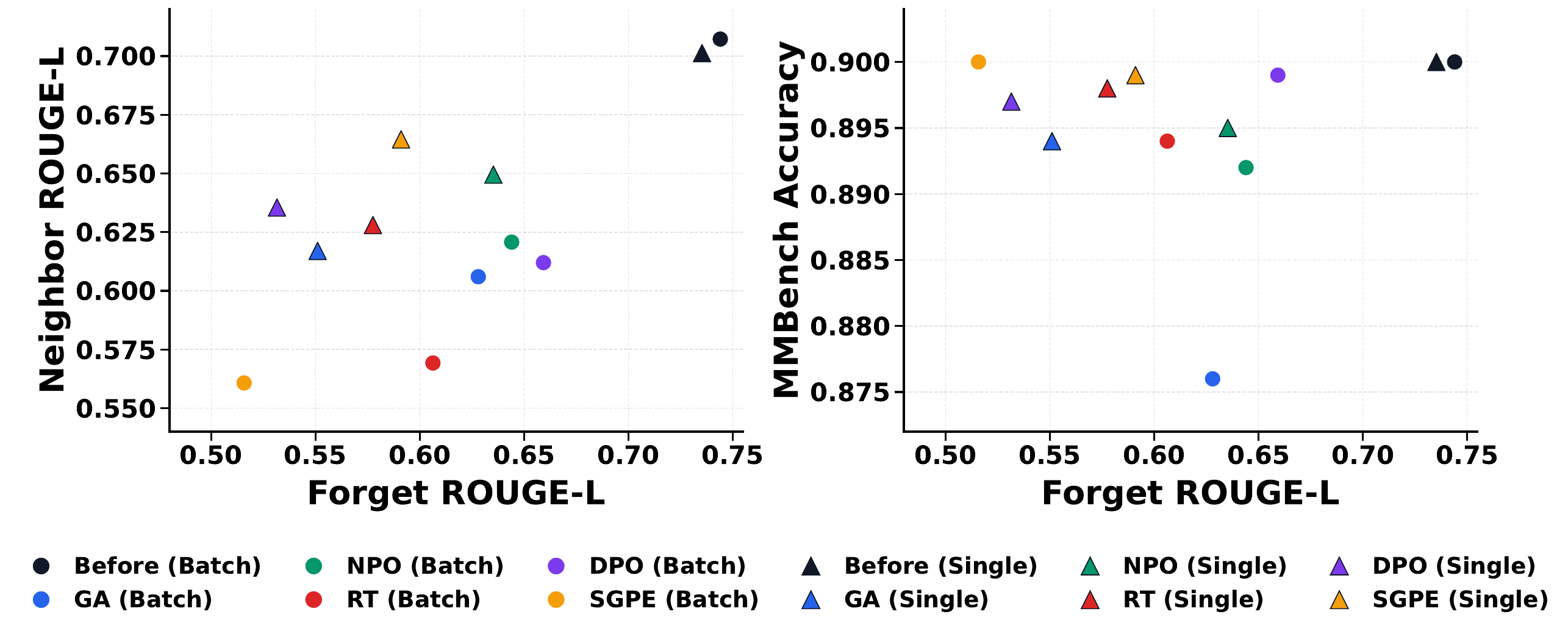}\\
\vspace{-1mm}
\caption{Trade-off between forgetting and knowledge preservation. 
The left plot shows Forget Set versus Neighbor Set performance, while the right plot shows Forget Set performance versus MMBench accuracy.}

\label{fig:trade-off}
\vspace{-3mm}
\end{figure}

\noindent\textbf{Trade-off between Forgetting, Locality, and Utility.} Fig.~\ref{fig:trade-off} compares forgetting performance, locality preservation, and general multimodal utility across different methods. The results reveal different forgetting-locality-utility trade-offs: methods with stronger forgetting capability often introduce larger collateral effects, whereas methods preserving utility may achieve weaker target suppression. SGPE provides a favorable trade-off among target suppression, locality preservation, and general multimodal capability.

\begin{figure}[t]
\centering
\includegraphics[width=1.0\linewidth]{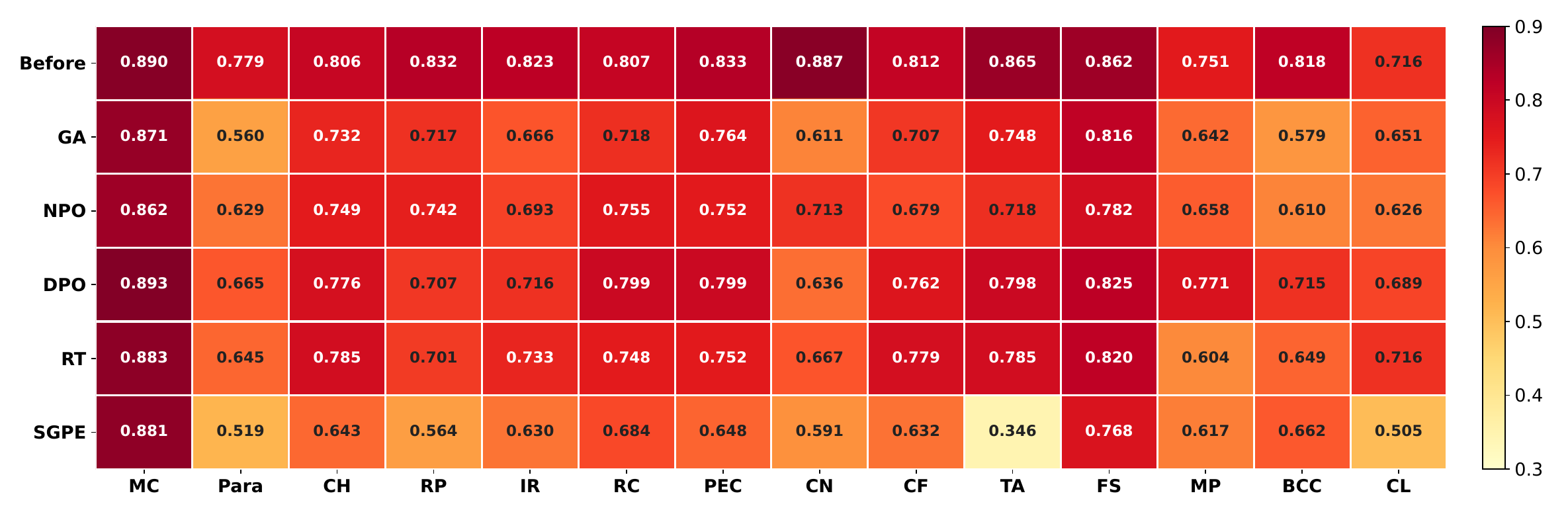}\\
\vspace{-1mm}
\caption{The heatmap reports ROUGE-L scores under 14 attack strategies, where lower values indicate stronger resistance to target knowledge reactivation. Attack abbreviations are defined in the Appendix.}

\label{fig:AA}
\vspace{-2mm}
\end{figure}

\noindent\textbf{Adversarial Robustness across Attack Styles.} While the main results evaluate AA probes globally, Fig.~\ref{fig:AA} further analyzes robustness across 14 adversarial attack styles. Lower ROUGE-L scores indicate weaker target knowledge recovery after unlearning. The results show that different attack strategies exhibit varying effectiveness in eliciting residual target knowledge, highlighting the necessity of adversarial evaluation beyond standard probes.

\setlength{\tabcolsep}{0.5pt}
\begin{table}[!t]
\caption{Ablation study of different SGPE components.}
\label{tab:ablation}
\renewcommand{\arraystretch}{1}
    \centering
\scriptsize
\begin{tabular}{l|cccc|cccc}
       \hlineB{2.5}
       \multirow{2}{*}{\textbf{Variants}} & \multicolumn{4}{c|}{\textbf{Forget Set $\downarrow$}} & \multicolumn{4}{c}{\textbf{Neighbor Set $\uparrow$}} \\
       \cline{2-9}
        & \textbf{QA (T)} & \textbf{QA (V)}& \textbf{AA (T)} & \textbf{AA (V)}& \textbf{QA (T)} & \textbf{QA (V)}&\textbf{AA (T)} & \textbf{AA (V)}\\
\hlineB{2}

SGPE
& 0.399&0.368&0.662&0.634&0.578&0.346&0.743&0.576\\

w/o Displacement Anchor
& 0.445&0.473&0.713&0.704&0.611&0.420&0.759&0.623\\

w/o Protected Projection
& 0.392&0.378&0.668&0.639&0.586&0.473&0.716&0.552\\

w/o Locality Gate
& 0.394&0.362&0.655&0.628&0.524&0.324&0.661&0.493\\

\hlineB{2.5}
\end{tabular}
\vspace{-3mm}
\end{table}

\noindent\textbf{Ablation Study.} We ablate three key components of SGPE: displacement anchor, protected projection, and locality gate. Removing the displacement anchor degrades forgetting performance, indicating the importance of providing a controlled forgetting direction. As shown in Tab.~\ref{tab:ablation}, removing protected projection leads to larger Neighbor Set degradation, suggesting that the orthogonal constraint helps preserve unrelated knowledge. Without the locality gate, the model achieves slightly stronger forgetting but suffers from increased collateral changes, highlighting the benefit of input-conditioned edit activation.

\noindent\textbf{Case Study.} We present qualitative examples of MLLM outputs before and after unlearning. As shown in Appendix, different methods exhibit distinct response behaviors: GA and NPO may produce unstable or hallucinated responses in some cases, while RT often encourages refusal-style outputs. In comparison, SGPE tends to produce more localized changes while maintaining responses to non-target queries.

\section{Conclusion}

In this work, we introduce PRMU, a benchmark for evaluating corpus-free multimodal unlearning under realistic person-centric unlearning scenarios. PRMU reveals that existing unlearning methods, although able to suppress target-related knowledge responses, often suffer from locality degradation and remain vulnerable to multimodal knowledge reactivation. We further provide SGPE, a lightweight corpus-free unlearning baseline that combines knowledge displacement, protected projection editing, and locality-aware multimodal control. Extensive experiments demonstrate that SGPE provides a competitive trade-off between forgetting and knowledge preservation under the PRMU setting. We hope PRMU can facilitate future research toward more realistic, scalable, and robust multimodal unlearning.

\bibliography{aaai2027}


\end{document}